\documentclass[runningheads,a4paper]{llncs}

\usepackage{amssymb}
\usepackage{amsmath}
\usepackage[compatibility=false,font={small}]{caption}
\usepackage{subcaption}
\usepackage{booktabs}
\usepackage{algpseudocode}
\usepackage{algorithm}
\usepackage[inline]{enumitem}
\usepackage{multirow}

\usepackage{graphicx}

\usepackage[pagebackref=true,breaklinks=true,colorlinks,bookmarks=false]{hyperref}

\usepackage{url}
\urldef{\mails}\path|{tommaso.cavallari, luigi.distefano}@unibo.it|

\begin{document}

\mainmatter  % start of an individual contribution

%PSIVT2015 submission
\def\PSIVT2015SubNumber{57} %Insert your submission number here

\title{Volume--based Semantic Labeling with Signed Distance Functions}

\titlerunning{Volume--based Semantic Labeling with Signed Distance Functions}

\author{Tommaso Cavallari \and Luigi Di Stefano}
%\author{Alfred Hofmann%
%\thanks{Please note that the LNCS Editorial assumes that all authors have used
%the western naming convention, with given names preceding surnames. This determines
%the structure of the names in the running heads and the author index.}%
%\and Ursula Barth\and Ingrid Haas\and Frank Holzwarth\and\\
%Anna Kramer\and Leonie Kunz\and Christine Rei\ss\and\\
%Nicole Sator\and Erika Siebert-Cole\and Peter Stra\ss er}
%
\authorrunning{Volume--based Semantic Labeling with Signed Distance Functions}

\institute{Department of Computer Science and Engineering\\
University of Bologna, Bologna, Italy\\
\mails}


\toctitle{Volume--based Semantic Labeling with Signed Distance Functions}
\tocauthor{Tommaso Cavallari, Luigi Di Stefano}
\maketitle

%%%%%%%%% ABSTRACT
\begin{abstract}
Research works on the two topics of Semantic Segmentation and SLAM (Simultaneous Localization and Mapping) have been following separate tracks. Here, we link them quite tightly by delineating a category label fusion technique that allows for embedding semantic information into the dense map created by a volume-based SLAM algorithm such as KinectFusion. Accordingly, our approach is the first to provide a semantically labeled dense reconstruction of the environment from a stream of RGB-D images. We validate our proposal using a publicly available semantically annotated RGB-D dataset and a) employing ground truth labels, b) corrupting such annotations with synthetic noise, c) deploying a state of the art semantic segmentation algorithm based on Convolutional Neural Networks.
\end{abstract}

%%%%%%%%% BODY TEXT
\section{Introduction}
\label{sec:introduction}

In the last years the Computer Vision community renewed its interest in the task of Simultaneous Localization and Mapping by leveraging on RGB-D information. This research trend has been fostered  by the development of ever cheaper sensors as well as by the more and more ubiquitous presence of smart mobile platforms, possibly having such sensors on board.
Many works tackled issues related to reliable camera tracking, accurate  mapping, scalable world representation, efficient sensor relocalization, loop closure detection,  map optimization. A major breakthrough in the realm of RGB-D SLAM was achieved by the KinectFusion algorithm~\cite{Newcombe2011}, which firstly demonstrated real-time and accurate dense surface mapping and camera tracking. 

On separate tracks, researchers working on object detection and semantic segmentation proposed many interesting techniques to extract high-level knowledge from images by recognition of object instances or categories and subsequent region labeling.
Especially thanks to the recent developments in the field of deep convolutional neural networks, year after year, new benchmark-beating algorithms are proposed that enable to quickly process raw images and extract from them valuable semantic information.

\begin{figure}[htb]
	\centering
	\begin{subfigure}{0.48\linewidth}
		\includegraphics[width=\textwidth]{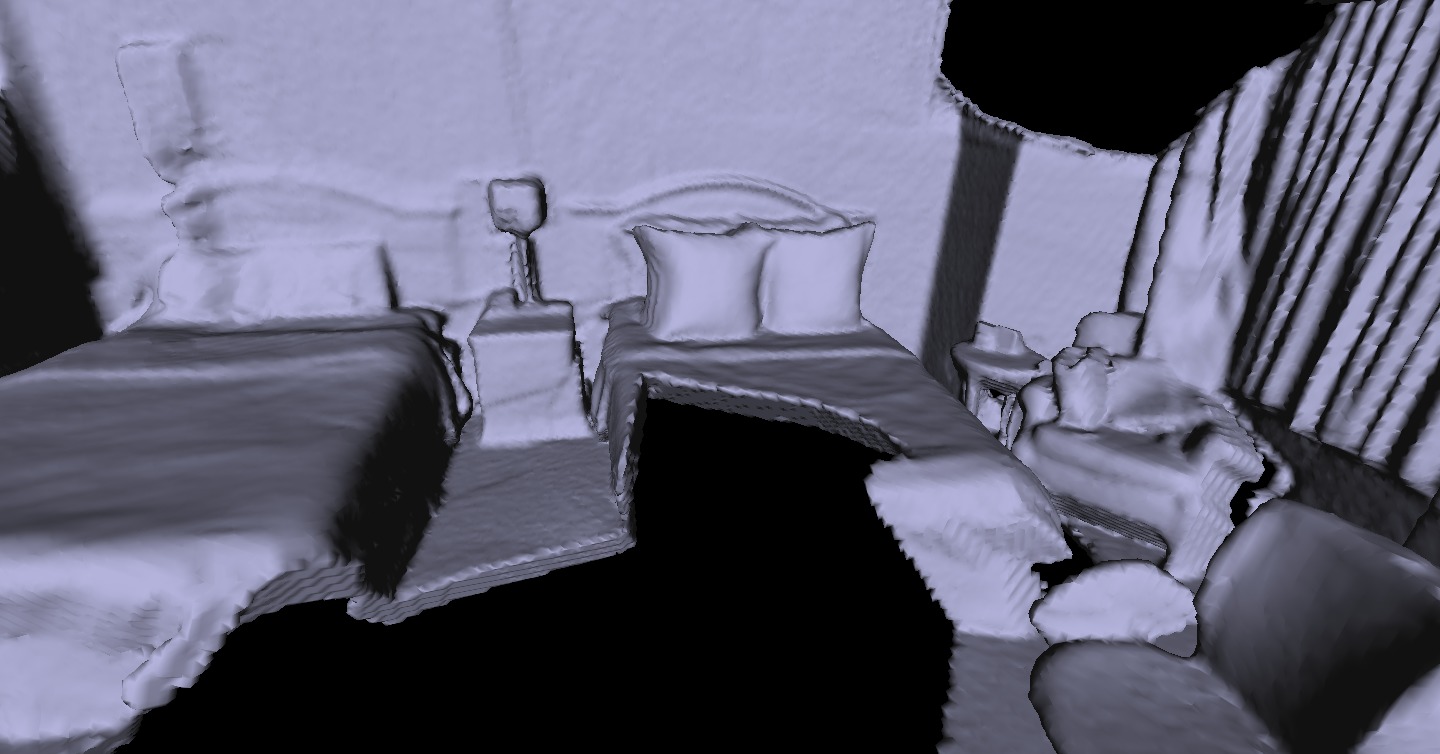}
	\end{subfigure}
	~
	\begin{subfigure}{0.48\linewidth}
		\includegraphics[width=\textwidth]{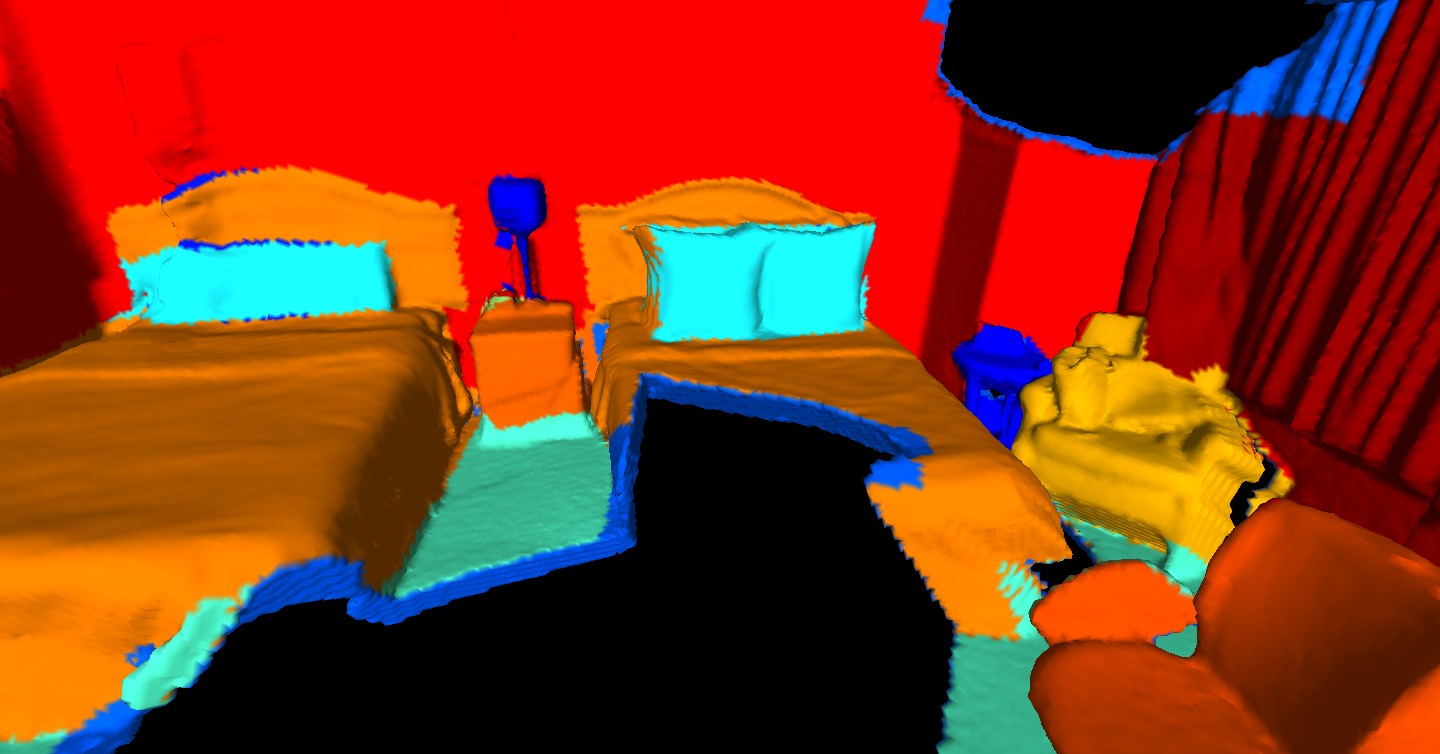}
	\end{subfigure}
	\caption{The left picture shows the standard KinectFusion output. The right picture illustrates the type of output delivered by our technique: a fully labeled dense reconstruction where each surface element is assigned a category tag.}
	\label{fig:teaser}
\end{figure}

However,  just a few works have tried to draw a bridge between the two aforementioned fields, though
we believe that both research areas could benefit significantly from tighter integration. Indeed, a SLAM process may be improved by deploying high-level knowledge on the type of objects encountered while the moving agent explores the environment, whereas object detection and semantic labeling techniques could be ameliorated by deploying multiple views from tracked sensor poses.

In this article we propose a technique capable to obtain incrementally a dense semantic labeling of the environment from a stream of RGB-D images while performing tracking and mapping \`{a} la KinectFusion~\cite{Newcombe2011}.  
Therefore, differently from the map concerned only with the 3D shape of the surfaces present in the environment yielded by a typical SLAM algorithm such as KinectFusion, our technique additionally provides a fully labeled map that embodies the information on \emph{what} kind of object  (e.g. a wall, chair, bed, pillow, furniture..)  each reconstructed surface element belongs to.  
A view from one of such dense semantic maps is reported in~\autoref{fig:teaser}, with each color representing a different category label.

The rest of this article is organized as follows: \autoref{sec:related_works} discusses briefly some of the most relevant works aimed at connecting semantic perception and SLAM; \autoref{sec:method_description} describes first the camera tracking and mapping method employed in our work and, subsequently, illustrates our proposal concerning the integration of a semantic labeling algorithm's output within the SLAM framework; finally, \autoref{sec:evaluation}  shows how the proposed volume-based semantic labeling technique behaves when feeding it with 
\begin{enumerate*}[label=\itshape\alph*\upshape)]
    \item ``correct'', manually annotated, labels,
    \item labels corrupted by synthetic noise,
    \item ``real'' labels obtained by a state of the art semantic segmentation algorithm.
\end{enumerate*}

%-------------------------------------------------------------------------
\section{Related works}
\label{sec:related_works}
As mentioned in the previous section, embedding of semantic informations into SLAM algorithms was addressed by just a few works.
A relevant early proposal in this field is the work by Castle et~al.~\cite{Castle2010}, where location of planar objects detected by SIFT features are incorporated into a SLAM algorithm based on Extended Kalman Filtering.
Later, Civera et~al.~\cite{Civera2011}, extended the previous approach to account for non-planar objects. 
Bao et~al.~\cite{Bao2011} proposed the idea of ``Semantic Structure from Motion'' to jointly perform the object recognition and SLAM tasks; in their research, tough, they process entire image sequences offline and perform a global optimization on the resulting environmental map.
All such approaches, also, do not employ RGB-D informations, relying instead on the processing of several images to estimate the 3D world structure.

On the converse, works exploiting the availability 3D information throughout the entire pipeline are those by Fioraio et~al.~\cite{Fioraio2013}, Salas-Moreno et~al.~\cite{Salas-Moreno2013} and Xiao et~al.~\cite{Xiao2013}.
Fioraio proposes a keyframe-based SLAM algorithm where detected objects are inserted as additional constraints in the bundle adjustment process used to estimate camera poses.
The work by Salas-Moreno relies instead on a pipeline where only detected objects are used to estimate sensor location by rendering a synthetic view of their placement and aligning the real depth image to such view through the ICP algorithm~\cite{Rusinkiewicz2001}.
Xiao introduces a semantically annotated dataset; while not the main focus of his work, semantic informations on the object location are used during the bundle adjustment process to better constrain the generated reconstruction of the environment.
In their work they show a full ``semantic loop'' where bounding boxes for objects manually labeled in a subset of frames are used to improve the world map; in turn this allows to propagate the labels to previously unlabeled frames in order to reduce the effort needed by the user to annotate the entire sequences.

Therefore, to the best of our knowledge, no previous work has attempted to bridge the gap between semantic segmentation and SLAM in order to achieve a dense semantic reconstruction of the environment from a moving visual sensor.

%------------------------------------------------------------------------
\section{Description of the method}
\label{sec:method_description}
To obtain a densely labeled map of the environment captured by the sensor, we adopt a volume-based approach. Similarly to  KinectFusion  \cite{Newcombe2011}, the map is represented
by a Signed Distance Function~\cite{Curless1996}, but, peculiarly, we also provide each voxel with a label that specifies  
the type of object appearing in that spatial location together with an indication of the confidence on the assigned label. 

To update the information stored into the volume by integrating new measurements, we need to track the RGB-D sensor as it moves within the environment.
In KinectFusion \cite{Newcombe2011} camera tracking is performed by ICP-based alignment between the surface associated with the current depth image and that extracted from the TSDF. 
Later, Bylow  et~al.~\cite{Bylow2013a} and Canelhas et~al.~\cite{Canelhas2013} proposed to track the camera by direct alignment of the current depth image to the mapped environment encoded into the TSDF 
as the zero-level isosurface. 
This newer approach has been proven to be faster and more accurate than the original KinectFusion tracker.
In our work, we decided to employ the aforementioned direct camera-tracking method on such considerations of speed and accuracy.
More precisely, our code has been obtained by properly modifying a publicly available implementation of the standard KinectFusion algorithm\footnote{\url{https://github.com/Nerei/kinfu_remake}} in order to introduce both the direct camera tracking method as well as dense semantic labeling process.  

In the remainder of this section we will describe our proposed approach to achieve integration of semantic labels into the TSDF representation, while we refer the reader to the previously mentioned article by Bylow and colleagues~\cite{Bylow2013a} for details on the tracking algorithm's implementation.
 
\subsection{Labeled TSDF}
\label{sec:labeled_tsdf}
To obtain a densely labeled representation of the environment, we assume the RGB-D sensor output to be fed to a semantic segmentation algorithm.
Without lack of generality, the output of such an algorithm can be represented as a ``category'' map, i.e. a bitmap having the same resolution as the input image wherein each pixel is assigned a discrete label identifying its category or the lack thereof.
Moreover, we assume to be provided with a ``score'' map, where each value represents the confidence of the labeling algorithm in assigning a category to the corresponding pixel of the input image.
Different semantic segmentation algorithms may indeed produce their output in heterogeneous formats (e.g. per-pixel categories, labeled superpixels, 2D or 3D bounding boxes, 3D cluster of points, polygons\ldots) but it is typically possible to reconcile those into the aforementioned intermediate representation.
The reliance on such an algorithm-agnostic format may also allow us to exploit simultenously the output from diverse labeling techniques, either aimed at detection of different categories or in order to combine their predictions by fusing the score maps.
As not every semantic segmentation technique can be run in real time for every frame captured by the sensor, our proposed label storage and propagation technique also allows for robust integration of unlabeled frames. 

A typical TSDF volume
holds for each voxel the (truncated) distance of its center from the closest surface in the environment.
A weight (also truncated to a maximum value) is stored alongside the distance to compute a running average of the SDF value while tracking the sensor~\cite{Newcombe2011}.

To label an element of the volume, several approaches may be envisioned.
The most informative is to store, as an histogram, a probability density function representing the probability for a voxel to represent an object of a certain class.
Advantage of this approach is the possibility to properly label a multi-category voxel (such as a voxel spatially located between two or more objects), also, analogously to the trilinear interpolation of SDF values, one may interpolate between neighboring voxels to obtain a spatially continuous p.d.f. %controllare
Unfortunately, practical memory occupancy issues forbid us to rely on such an approach: each voxel already holds an SDF value and a weight, stored as half precision floating point numbers, i.e. 4 bytes of  memory storage.
With a 512$^3$ voxel grid this means a memory occupancy of 512 MB, with typical consumer GPU cards rarely providing more than 2-3 GB of total usable memory. 
An RGB triplet may also be stored for visualization purposes, this requiring 4 more bytes for each voxel\footnote{Due to memory alignment constraints it is not recommended to store only three bytes per triplet.} and doubling memory occupancy. 
Hence, by encoding the probability of each class as a half precision float number we could not store probabilities for more than 4-8 categories without filling up all the available GPU memory. 
Also, since the integration of a new frame into the voxel grid during the ``mapping'' step of the algorithm requires a visit to each voxel, the more categories one wishes to handle the slower turns out the entire tracking pipeline.

The above considerations lead us to store a single category per voxel, together with a ``score'' expressing the confidence on the accuracy of the assigned label.
Discrete labels are stored as unsigned short numbers while the score is once again an half precision float, which amounts to a total of 8 bytes per voxel including the already existing SDF data.
Accordingly, a 512$^3$ voxel grid requires 1 GB of GPU RAM regardless of the total number of handled categories.
Clearly, we lose information using such label encoding as we can no longer represent properly those voxels featuring more than one likely label.
Moreover, such a minimal representation mandates special care in implementing the volume update operation to insert new labeled data into the grid, in order to avoid situations where a voxel gets continuously switched between different categories.

\subsection{Volume update process}
\label{sec:volume_update}

Integration of the acquired depth values into the TSDF volume after camera pose estimation is a crucial operation for any volume-based tracking and mapping algorithm.
Typically, every voxel is visited in parallel on the GPU (as each is independent from others) and its 3D position is projected onto the depth map to select a pixel.
Such depth pixel stores the distance of the camera from the observed surface at the current frame (if available).
Analogously, the voxel stores the distance from its center to the closest surface.
The stored TSDF value then undergoes an update step consisting in the weighted average between the voxel's distance from the surface (obtained from the depth map) and the current TSDF value.
Weight is also increased, up to a maximum value.
Infinite weight growth is avoided to allow for temporal smoothing of the estimated surface distance: this approach allows older measurements to be forgotten after a certain number of volume updating steps. 

As mentioned, to store semantic information into each voxel, we add a discrete category label together with a floating point score expressing the confidence in the stored label. 
Hence, a running average approach such as that just described to update the map information cannot be used for the semantic labeling information as different categories cannot be directly confronted. 
We could store in each voxel the labels as we receive them (by projecting each 3D cell's coordinates onto the label bitmap and sampling the corresponding pixel), but this would be prone to errors as a mislabeled region would possibly overwrite  several correct voxel labels acquired in the past.
Additionally, not every pixel may be labeled, possibly entire frames when using a slow semantic segmentation technique which cannot be run on every input image.

Therefore, in this work  we propose an evidence weighting approach: each time an already labeled voxel is seen and associated to the currently stored category we increment its score.
If a labeled voxel in a subsequent frame is assigned to a different category (e.g. due to a labeling error or to being on the seam between two differently labeled regions), we decrement the associated score. Only when the score reaches a negative value we replace the stored category with the new one. 

As for the evidence increment/decrement weight applied to the score, we deploy the confidence of the semantic segmentation algorithm, as sampled from the input score map that we assume to be provided together with the labeling output itself. 
This choice naturally induces an hysteresis-like effect, protecting the consistency of the labels stored into the volume when areas of the input image are assigned to different categories in subsequent frames: typically a mislabeled region has associated a low score, such value then will not bring in enough evidence to change the category associated with a correctly identified area of the space.
Conversely, assuming the initial labeling of a region to be  wrong (therefore with a low confidence), a correct labeling from subsequent frames will easily be able to replace the initial erroneous tag. 
A possible pitfall becomes evident if, for any reason, the score associated to a wrong labeling result is very high but, as more frames are integrated into the TSDF volume, stored weights will increase above the maximum score the semantic segmentation algorithm is able to provide; when such a situation is reached, a single incorrect segmentation will not be able to adversely affect the volume contents.
An unlabeled area (or entire frame, without lack of generality) has no effect on the volume labeling process: each corresponding voxel will be left unchanged.

Similarly to the geometric integration approach, we clamp the maximum label score for a voxel  to allow for an easier change of category if suddenly a region of space is consistently tagged as a different object for several frames (e.g. in non static situations when an object is removed from the scene). Algorithm~\ref{alg:label_update} shows the pseudo-code for the proposed volume-based label updating process. 

\begin{algorithm}[htb]
    \caption{Pseudo-code of the label updating process}
    \label{alg:label_update}   
    \begin{algorithmic}
        \ForAll{voxels in the volume}
            \State $(i,j) \gets$ projection of the 3D voxel coordinates into the image plane
            \State $L_{in} \gets$ category associated to the pixel $(i,j)$
            \State $W_{in} \gets$ labeling score associated to the pixel $(i,j)$
            \State $L_{tsdf} \gets$ category associated to the current voxel
            \State $W_{tsdf} \gets$ labeling score associated to the current voxel
            \If{$L_{in} \notin (\text{unlabeled}, \text{background})$}
                \If{$L_{tsdf} = \text{unlabeled} \lor W_{tsdf} < 0$}
                    \State $L_{tsdf} \gets L_{in}$
                    \State $W_{tsdf} \gets W_{in}$
                \ElsIf{$L_{in} = L_{tsdf}$}
                    \State $W_{tsdf} \gets \min(W_{tsdf} + W_{in}, W_{clamp})$
                \Else
                    \State $W_{tsdf} \gets W_{tsdf} - W_{in}$
                \EndIf
            \EndIf
        \EndFor
    \end{algorithmic} 
\end{algorithm}
%------------------------------------------------------------------------
\section{Experimental evaluation}
\label{sec:evaluation}
To evaluate the proposed volume labeling approach we performed tests using different types of semantically segmented data.
Our tests deploy the video sequences included in the Sun3D dataset~\cite{Xiao2013}.
On their website, Xiao et~al., provide multiple RGB-D video sequences captured using a Kinect sensor and depicting typical indoor environments such as hotel, conference rooms or lounge areas. 
Unique to this dataset is the presence of manually acquired accurate object annotations in the form of per-object polygons for multiple sequences. 
Each object is also given a unique name, which allows us to tell apart several instances of a same category (e.g. in a hotel room sequence we may have ``pillow 1'' and ``pillow 2'').

\begin{figure}[htb]
	\centering
    \includegraphics[width=0.8\linewidth]{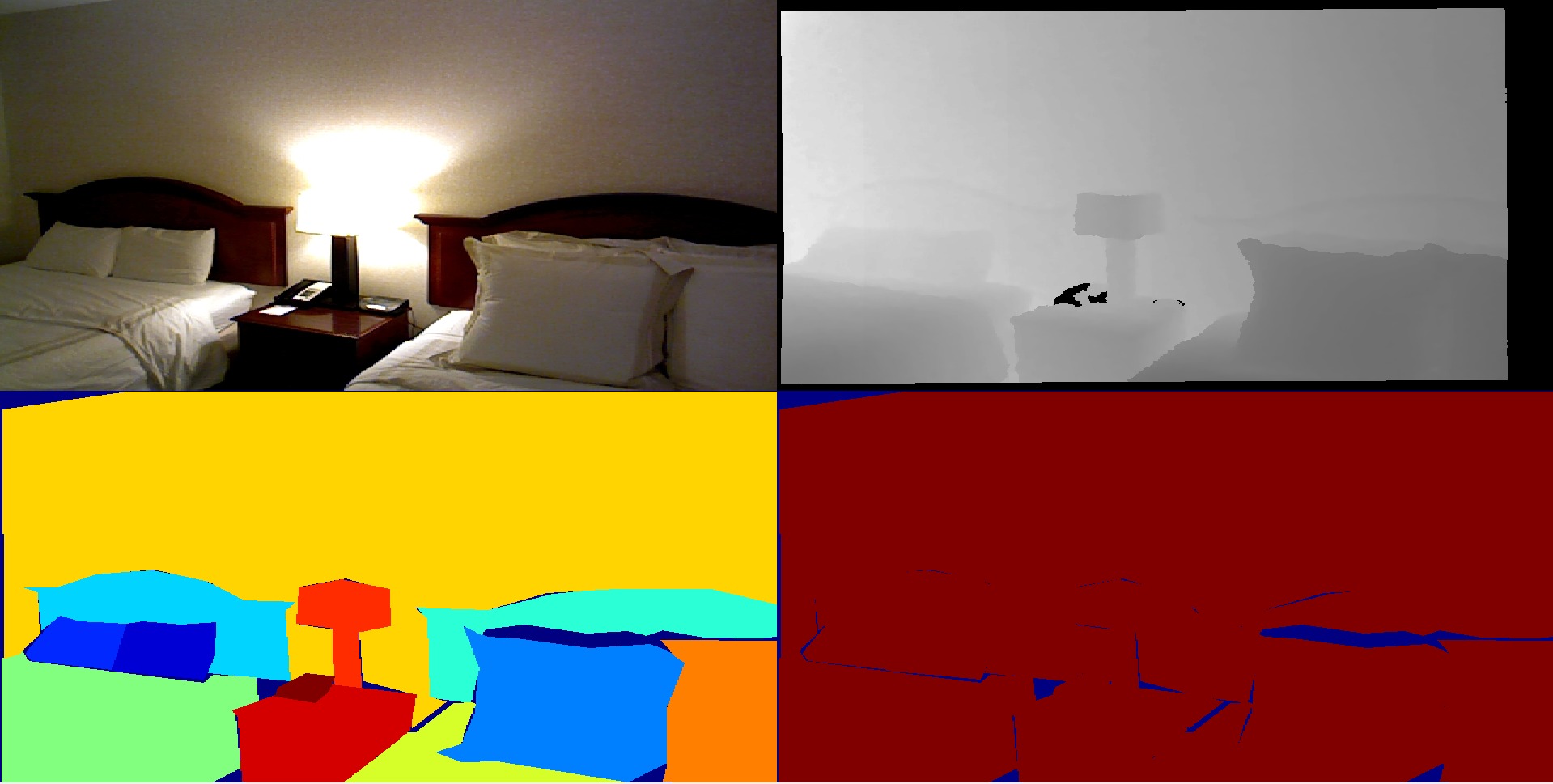}
    \caption{Labeled frame from a sequence of the Sun3D dataset. 
    From top left in clockwise order: RGB frame, depth frame, score map and category map. 
    Each map has been drawn in false colors to increase visibility (blue is the minimum value while red is the maximum).}
    \label{fig:labeled_frames}
\end{figure}

To parse the dataset's own object representation into our intermediate labeling format, described in \autoref{sec:labeled_tsdf}, we adopted the following approach:
\begin{description}
    \item[Category map] Each named object was given an increasing (and unique) integer identifier, afterwards, its bounding polygon was painted as a filled shape into our category bitmap. 
    Being the source data the result of a manual annotation process, partial overlap of the object polygons has not been a concern.
    \item[Score map] Annotated shapes have been manually defined, we therefore consider the \emph{labeling algorithm}'s confidence maximal.
    Similarly to the category bitmap, we draw each object's bounding polygon onto the score map and fill it with the floating point value $1.0$.
\end{description}
\autoref{fig:labeled_frames} shows a frame from the hotel room sequence contained in the aforementioned dataset, we see that each object is correctly labeled and their confidences are maximal due to the manual labeling process.

We will provide two kind of results, first by proving the robustness of the method under presence of synthetic noise in the labeler's output, subsequently, we will show densely labeled volumes for several sequences obtained using either ground truth labeling data or the semantic segmentation produced by a state of the art algorithm and we will evaluate the capability of the proposed fusion technique to reduce the number of erroneously labeled areas in the reconstructed volumes.

%------------------------------------------------------------------------
\subsection{Robustness to synthetic label noise}

\begin{figure}[htb]
    \centering
    \includegraphics[trim={0 0.25cm 0 1cm},width=\linewidth]{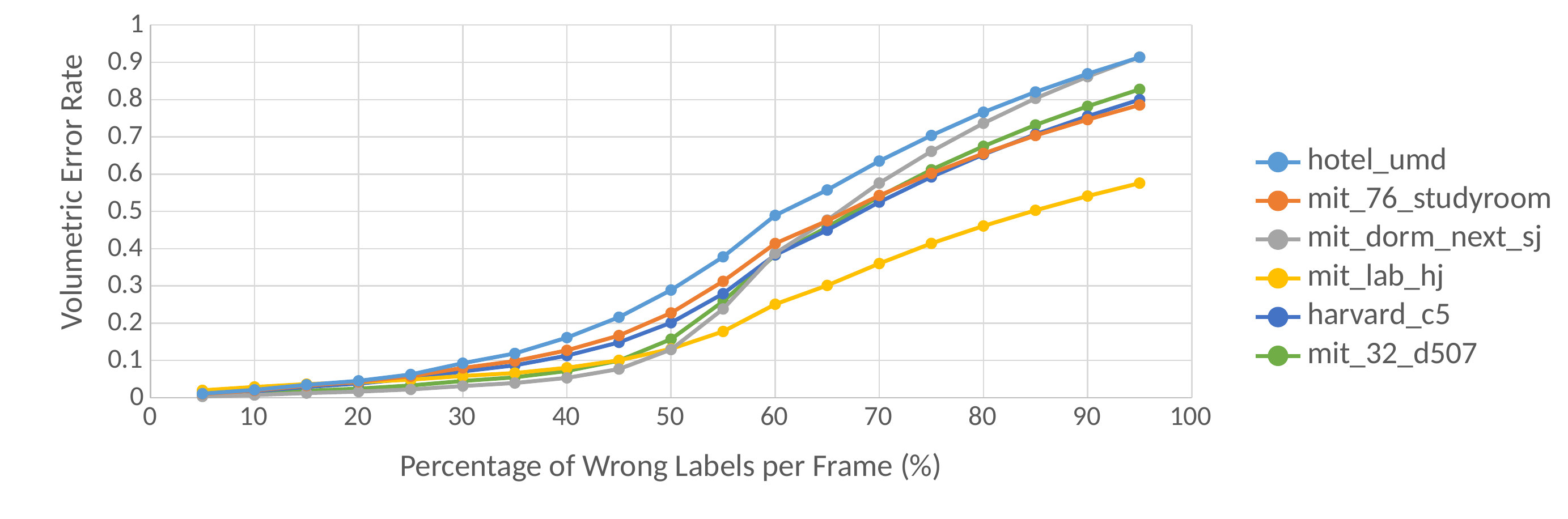}
    \caption{Volumetric labeling error rate for several sequences of the Sun3D dataset when synthetic noise is added to the ground-truth category labels.}
    \label{fig:eval_noise}
\end{figure}

% magari rifare i test aggiungendo noise anzichè sui pixel sui poligoni interi
To investigate on the robustness of the proposed volumetric label integration process with respect to per-pixel semantic segmentation errors, for all the considered sequences, we corrupted the ground-truth category map associated with each frame by employing synthetically generated white noise.  
Then, we compared the resulting labeled volume to the reference volume obtained by executing the label fusion process based on noiseless ground-truth category maps. In particular, considering only those labels assigned to voxels representing a surface element (i.e. the zero-level isosurface of the TSDF), we compute the volumetric labeling error rate, i.e. the fraction of misclassified surface voxels.
Our synthetic noise model is as follows.
We sample pixels from the category map with a certain probability so to switch their correct label to wrong ones uniformly selected from the total pool of labels present in the current sequence.
We also assign maximum confidence to switched labels.

\begin{figure*}
    \centering
    \begin{subfigure}[b]{0.2\linewidth}
        \includegraphics[width=\textwidth]{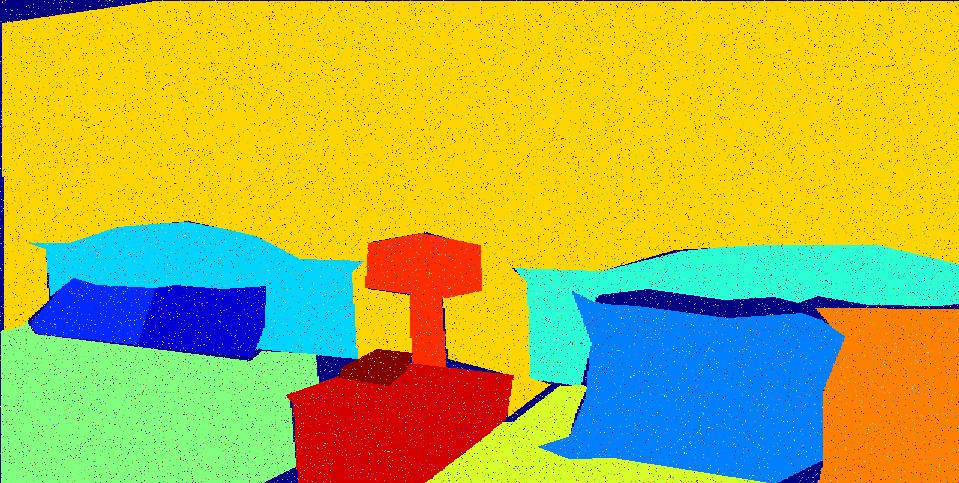}
        \caption{Noise: 5\%}
        \label{fig:eval_noisy_map_5}
    \end{subfigure}
    ~
    \begin{subfigure}[b]{0.2\linewidth}
        \includegraphics[width=\textwidth]{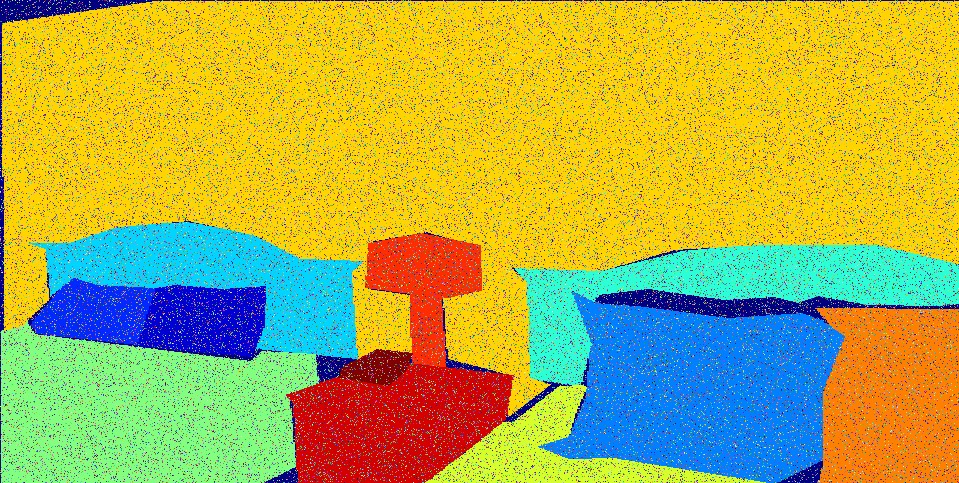}
        \caption{Noise: 15\%}
        \label{fig:eval_noisy_map_15}
    \end{subfigure}
    ~
    \begin{subfigure}[b]{0.2\linewidth}
        \includegraphics[width=\textwidth]{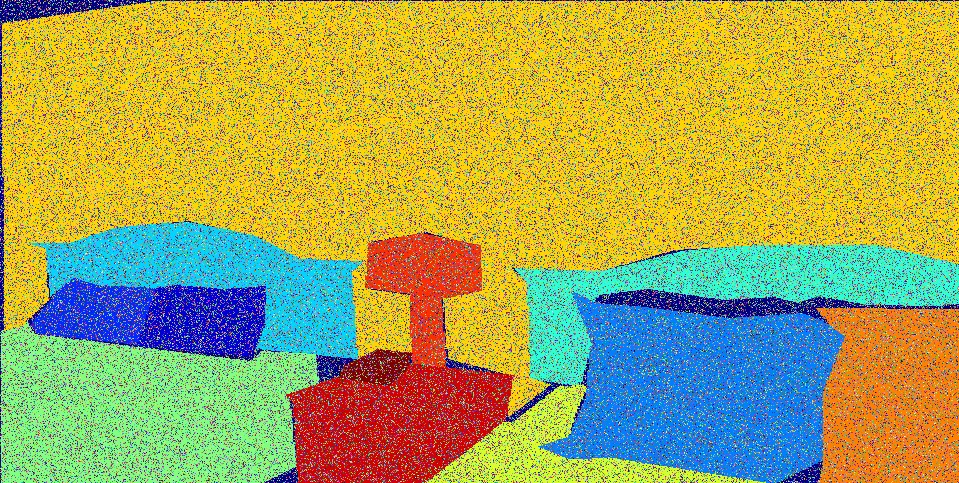}
        \caption{Noise: 30\%}
        \label{fig:eval_noisy_map_30}
    \end{subfigure}
    ~
    \begin{subfigure}[b]{0.2\linewidth}
        \includegraphics[width=\textwidth]{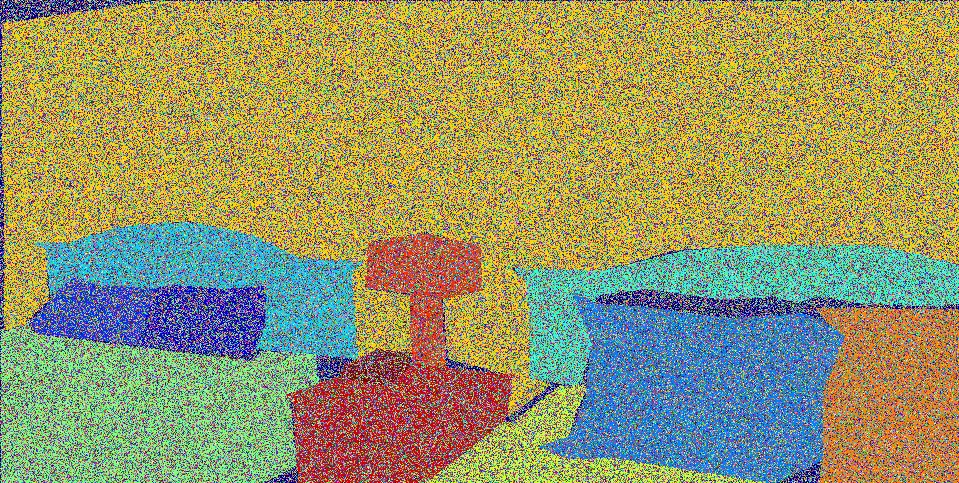}
        \caption{Noise: 60\%}
        \label{fig:eval_noisy_map_60}
    \end{subfigure}
    \caption{Sample category bitmaps when correct labels are corrupted by increasing amount of noise.}
    \label{fig:eval_noisy_map}
\end{figure*}
 
\autoref{fig:eval_noise} shows how, though the image labeler output is corrupted (as illustrated in \autoref{fig:eval_noisy_map}), thanks to the temporal label integration process, the final volume features a consistent labeling where each voxel is likely to have been correctly classified.
Even when the probability to corrupt a label is as high as 50\%, the proposed label integration can reduce the final volumetric error rate significantly, i.e. squeezing it down to less than 25\%  typically, to much less than 20\% quite often.
For more than 50\% of wrong labels per input image the error grows almost linearly with the noise level, the label fusion process still turning out beneficial in terms of noise attenuation: e.g, with as much as 70\% wrong labels per image the amount of misclassified surface voxels is typically less than 50\%.

\begin{figure}[htb]
    \centering
    \begin{subfigure}[b]{0.48\linewidth}
    	\includegraphics[width=\textwidth]{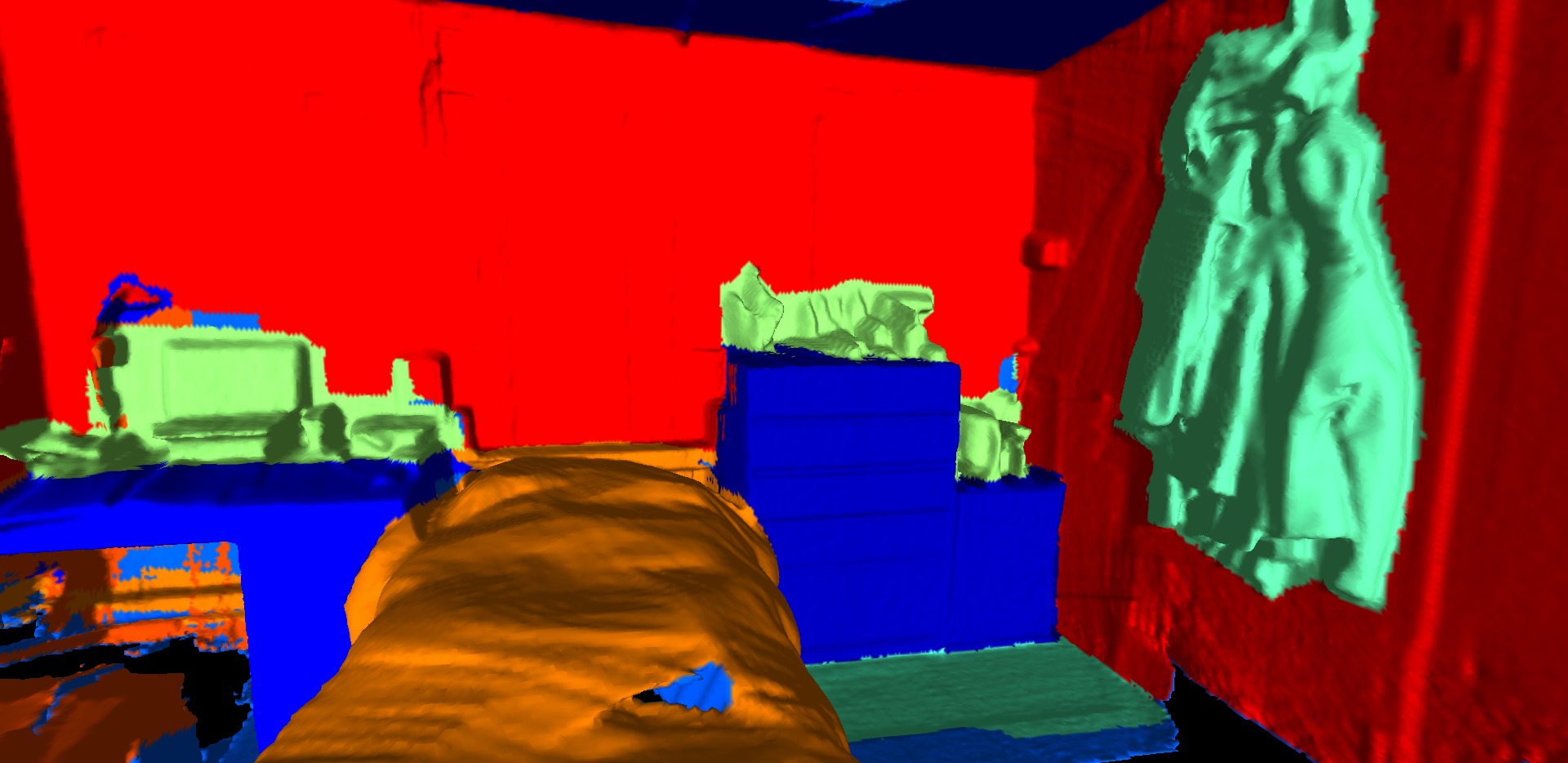}
        \caption{}
        \label{fig:eval_lab_vol_noiseless}
    \end{subfigure}
    ~
    \begin{subfigure}[b]{0.48\linewidth}
    	\includegraphics[width=\textwidth]{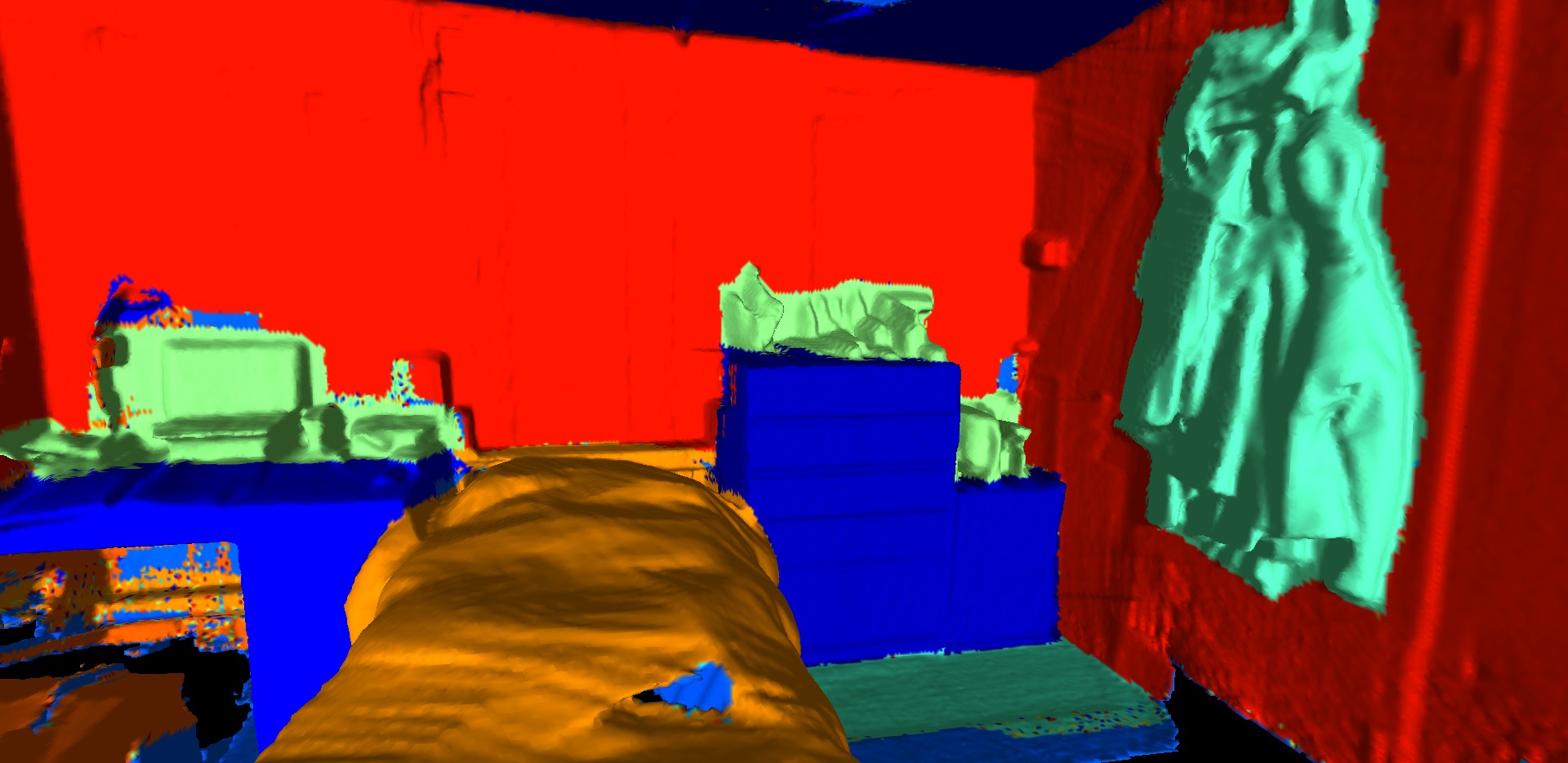}
        \caption{}
        \label{fig:eval_lab_vol_noisy}
    \end{subfigure}
    \caption{Semantically labeled reconstructions: each surface element is colored according to its category label. Left: reconstruction from noiseless per-pixel category maps. Right: reconstruction when  30\% of the input labels in each map are switched to wrong. Labeling errors are visible by zooming onto the desk area only.}
    \label{fig:eval_lab_vol}
\end{figure}

\autoref{fig:eval_lab_vol} depicts the semantic reconstruction of a portion of the environment explored through the mit\_dorm\_next\_sj sequence. 
It can be observed that the labeled surface represents accurately both the shape as well as the semantic of the objects present in the environment. The comparison between \autoref{fig:eval_lab_vol_noiseless} and \autoref{fig:eval_lab_vol_noisy} allows for assessing the effectiveness of the temporal label integration process: though as many as 30\% of the per-pixel labels in each frame are wrong, just a few errors are noticeable with respect to the semantic reconstruction based on perfect noiseless input data. Indeed, such errors are mostly concentrated in the desk area, where the  sensor did not linger for multiple frames and thus the evidence weighting process turned out less effective.

%------------------------------------------------------------------------
\subsection{Results in real settings}
%The previously shown results are entirely based on a ground truth composed of manually annotated images, the labeling accuracy was therefore maximal.
To evaluate the effectiveness of our technique when using a real semantic labeling algorithm, we used the recent Semantic Segmentation approach proposed by Long, Shelhamer and Darrell~\cite{Long2015}.
Such algorithm uses a Convolutional Neural Network to produce a per-pixel labeling of an input image. 
The authors made available several pretrained  networks\footnote{\url{https://github.com/BVLC/caffe/wiki/Model-Zoo\#fcn}} based on the open source Caffe deep learning framework~\cite{Jia2014}. 
We chose to employ the ``FCN-16s NYUDv2'' network due to similarities of the category set with the type of objects present in the Sun3D dataset. FCN-16s NYUDv2 processes RGB-D images and produces  per-pixel scores for 40 categories defined in~\cite{Gupta2013}.
Using the aforementioned algorithm to label each frame of the sequences, we fed our volume labeling pipeline with category maps where each pixel is assigned to the object class having the highest probability, storing then such values into the respective score maps.

\begin{figure*}[htb]
	\centering
	\begin{subfigure}[b]{0.45\linewidth}
		\includegraphics[width=\textwidth]{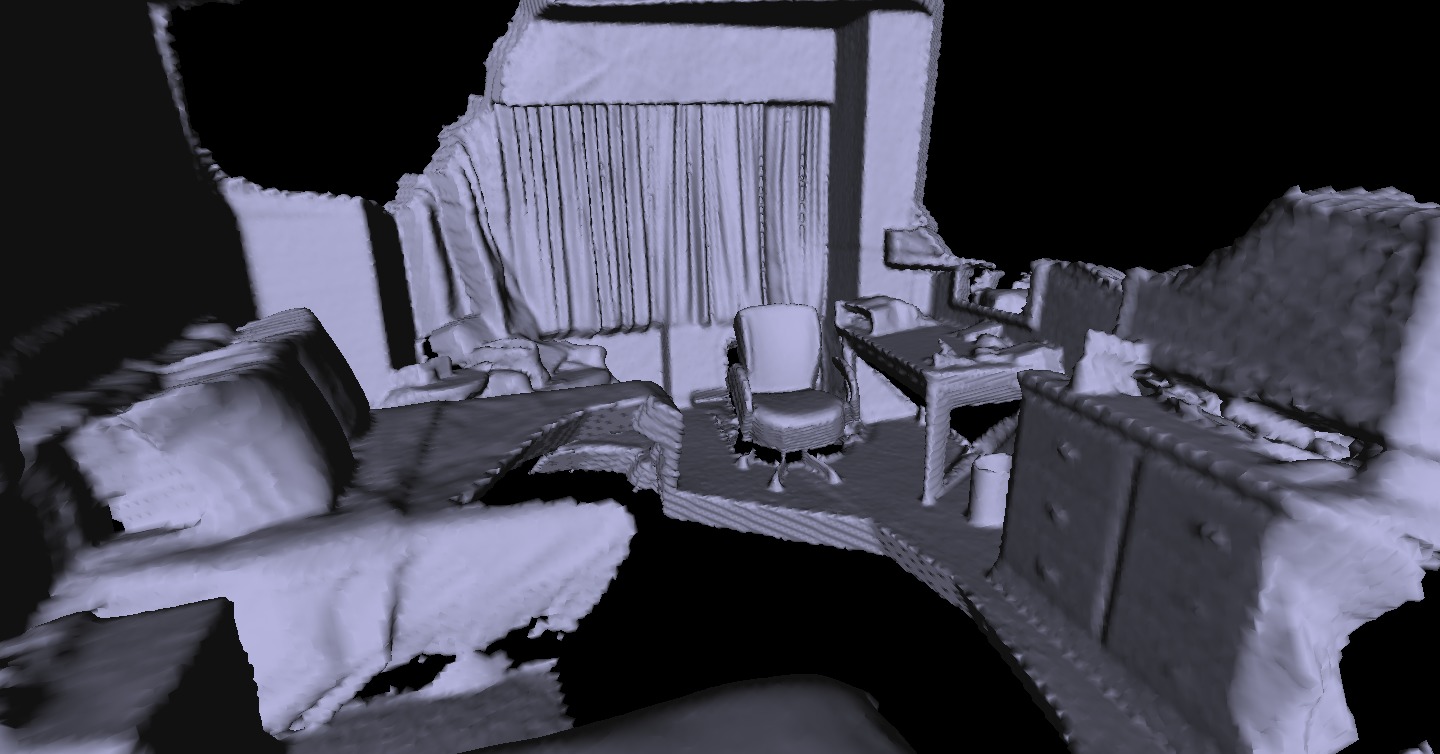}
	\end{subfigure}
	~
	\begin{subfigure}[b]{0.45\linewidth}
		\includegraphics[width=\textwidth]{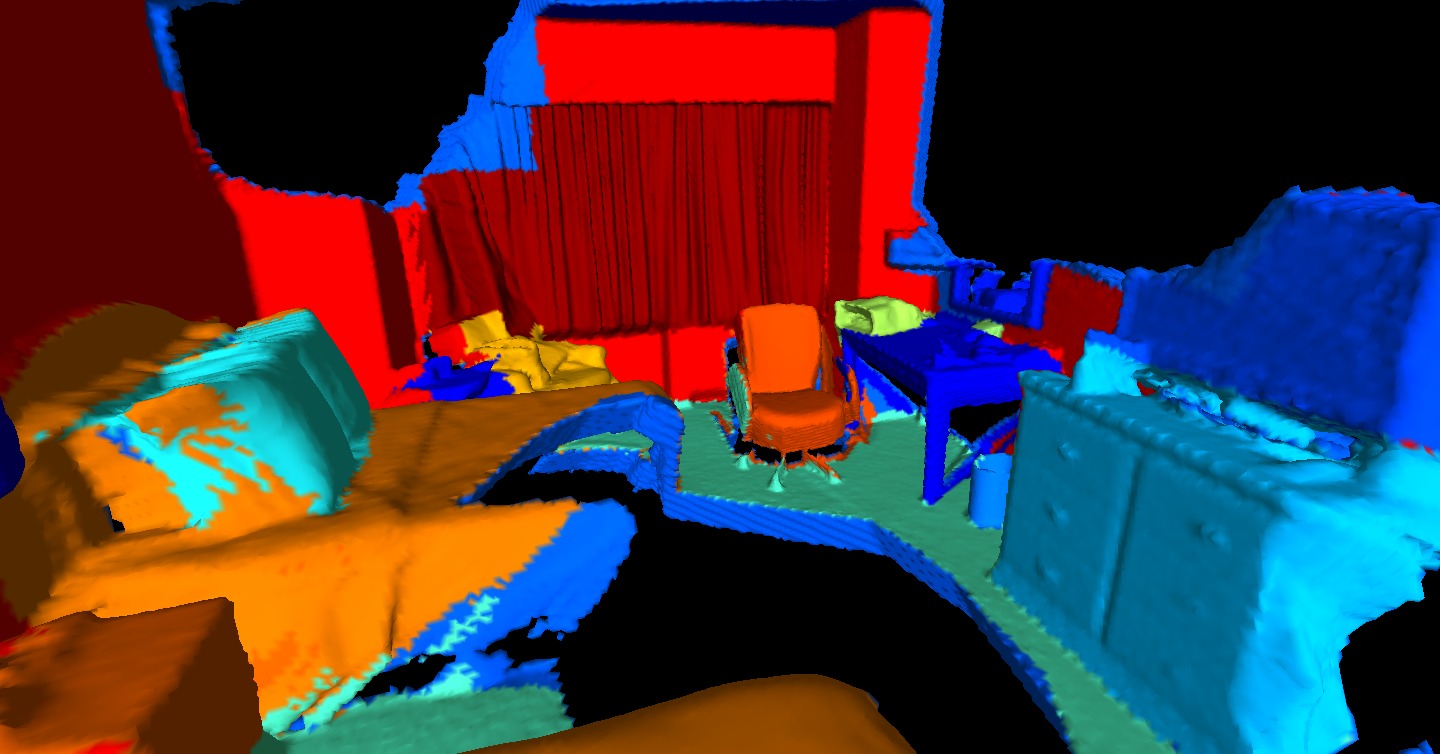}
	\end{subfigure}
	\\
	\begin{subfigure}[b]{0.45\linewidth}
		\includegraphics[width=\textwidth]{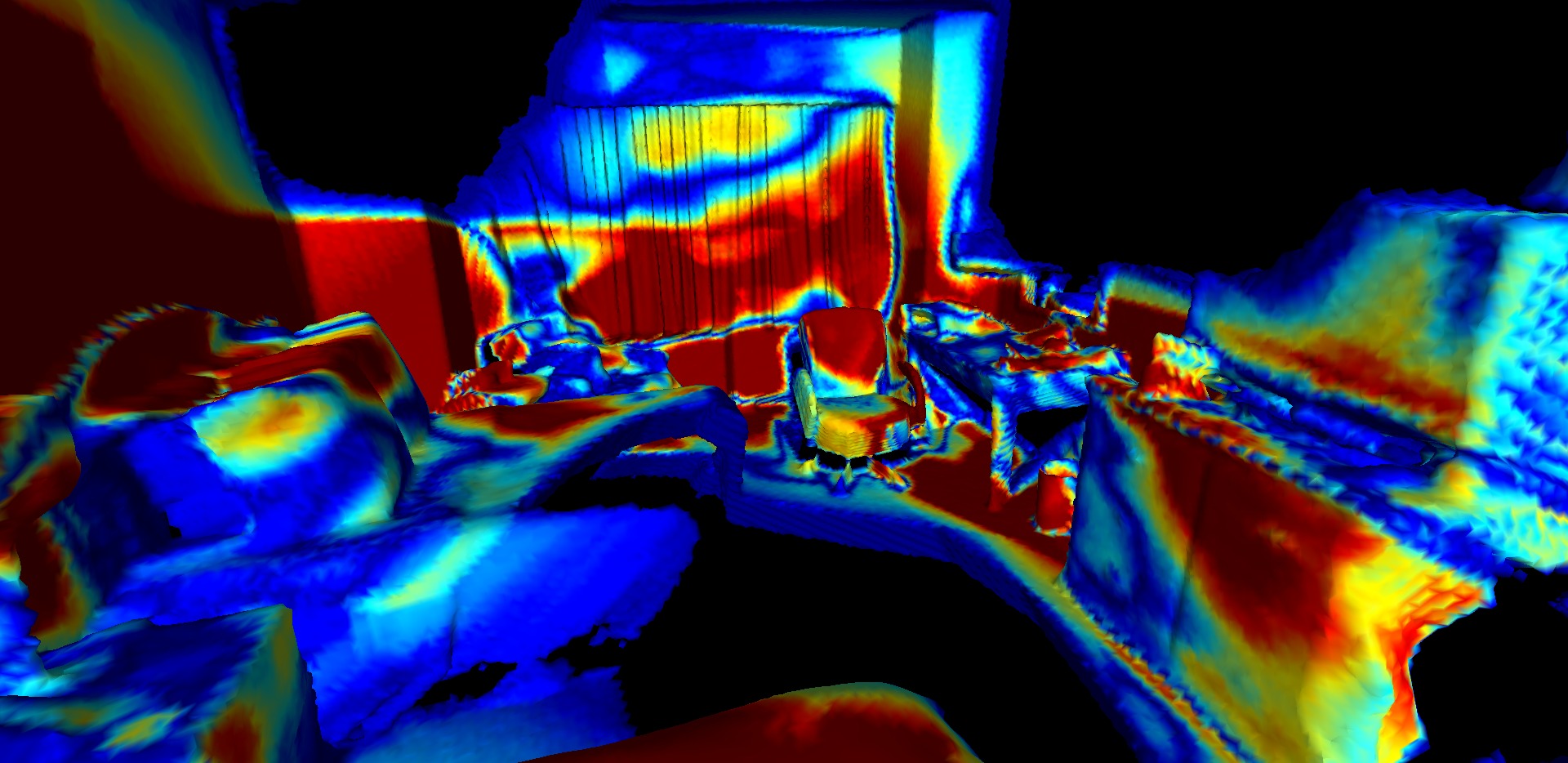}
	\end{subfigure}
	~
	\begin{subfigure}[b]{0.45\linewidth}
		\includegraphics[width=\textwidth]{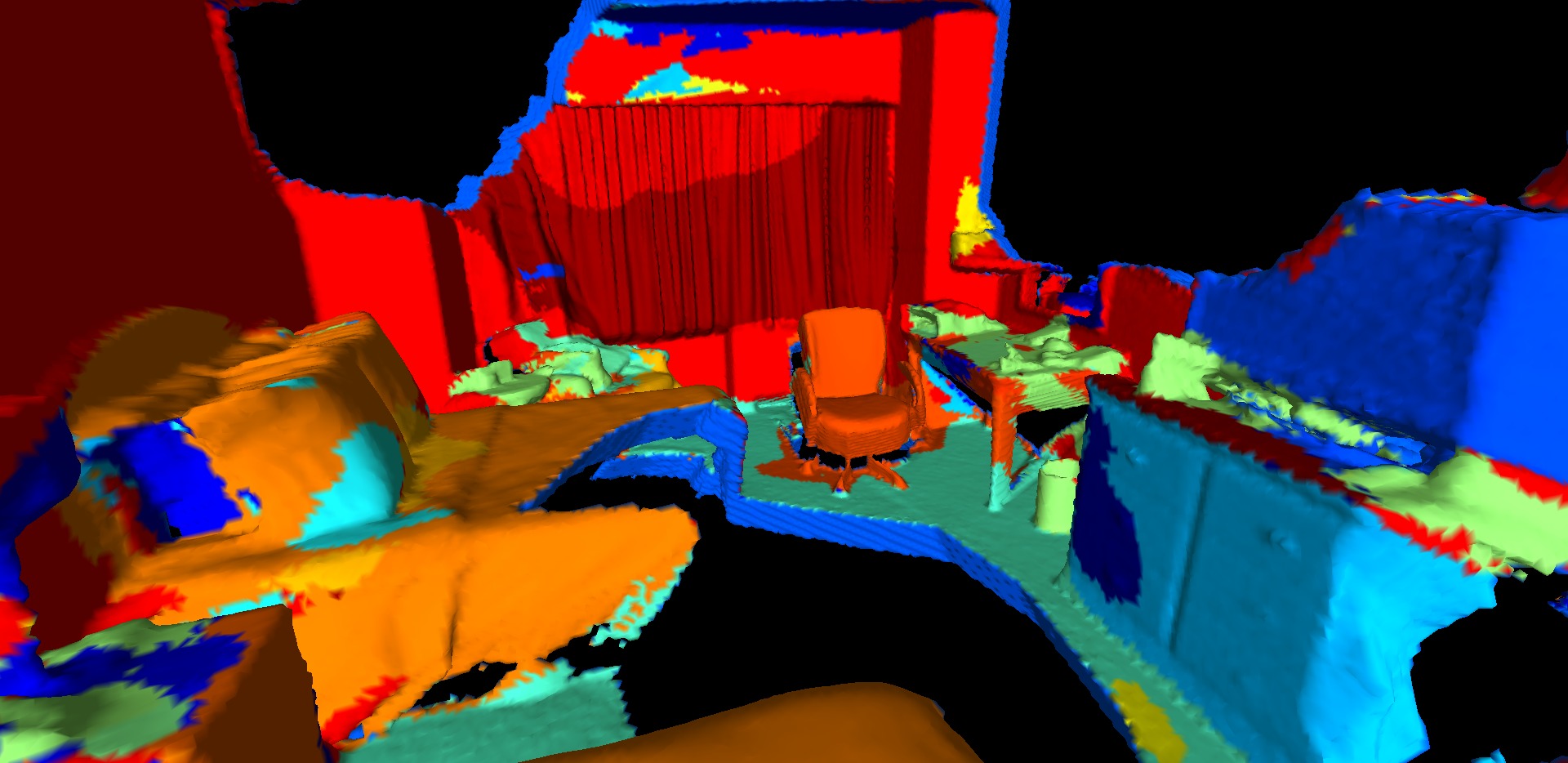}
	\end{subfigure}
	\caption{View from hotel\_umd sequence. From the top left, in clockwise order: standard KinectFusion output, semantically labeled view using manually annotated categories, semantically labeled view and related confidence map using the CNN.}
	\label{fig:eval_hotel_view_1}
\end{figure*}

\begin{figure*}[htb]
	\centering
	\begin{subfigure}[b]{0.45\linewidth}
		\includegraphics[width=\textwidth]{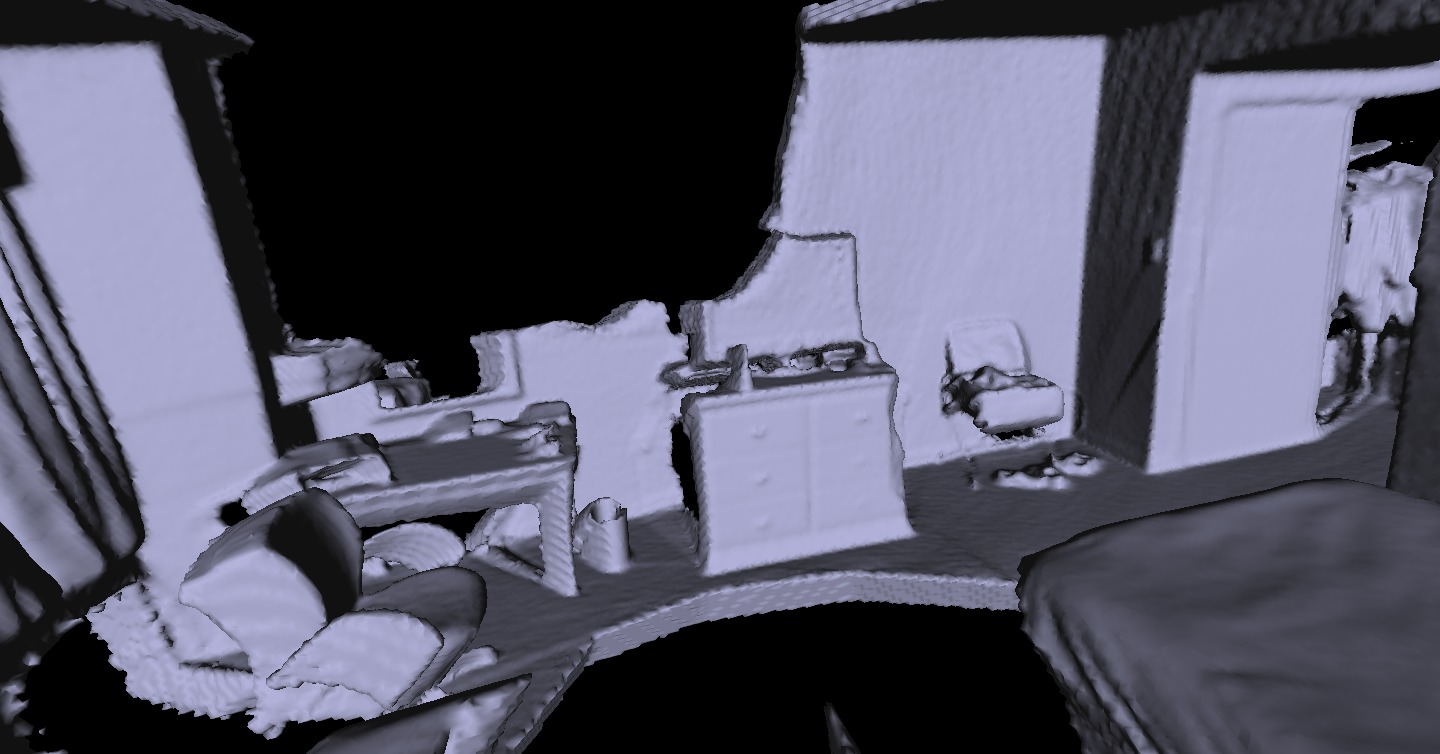}
	\end{subfigure}
	~
	\begin{subfigure}[b]{0.45\linewidth}
		\includegraphics[width=\textwidth]{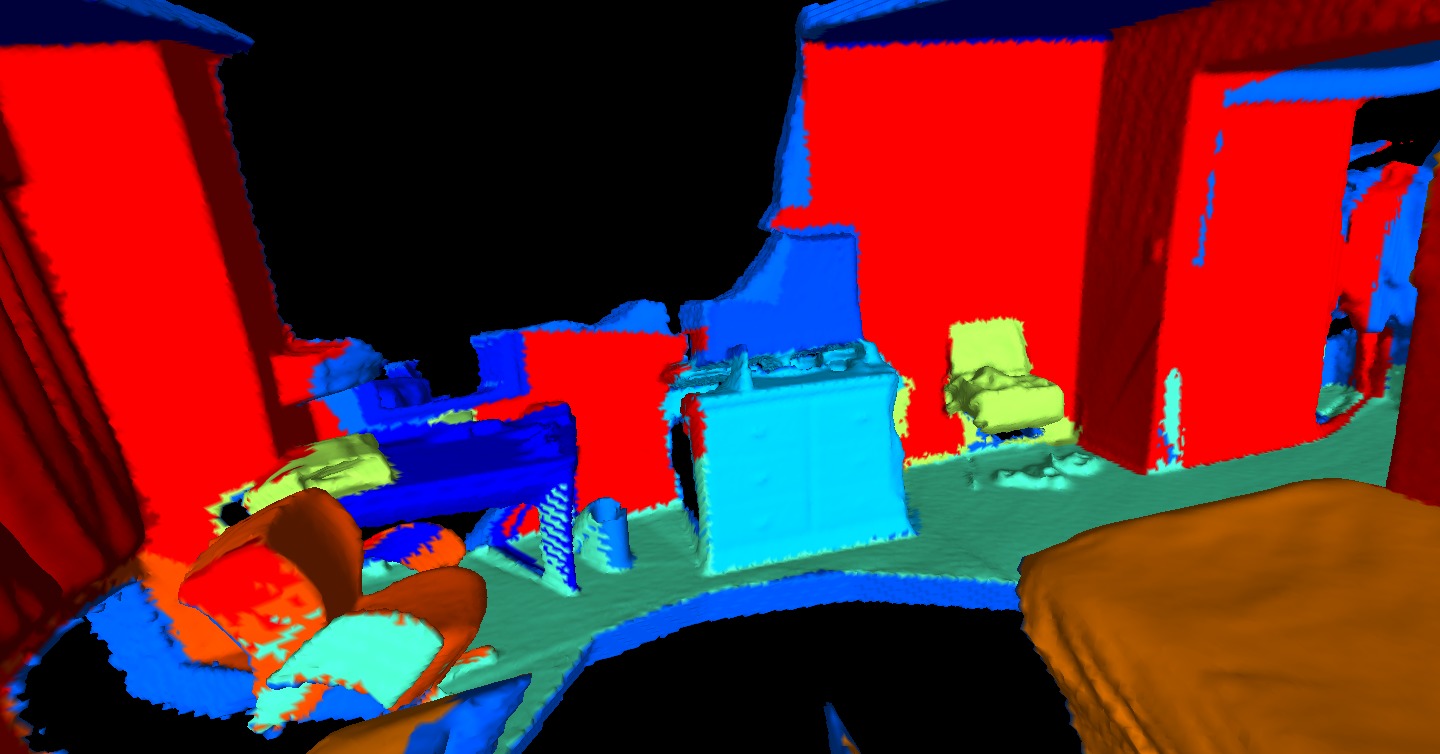}
	\end{subfigure}
	\\
	\begin{subfigure}[b]{0.45\linewidth}
		\includegraphics[width=\textwidth]{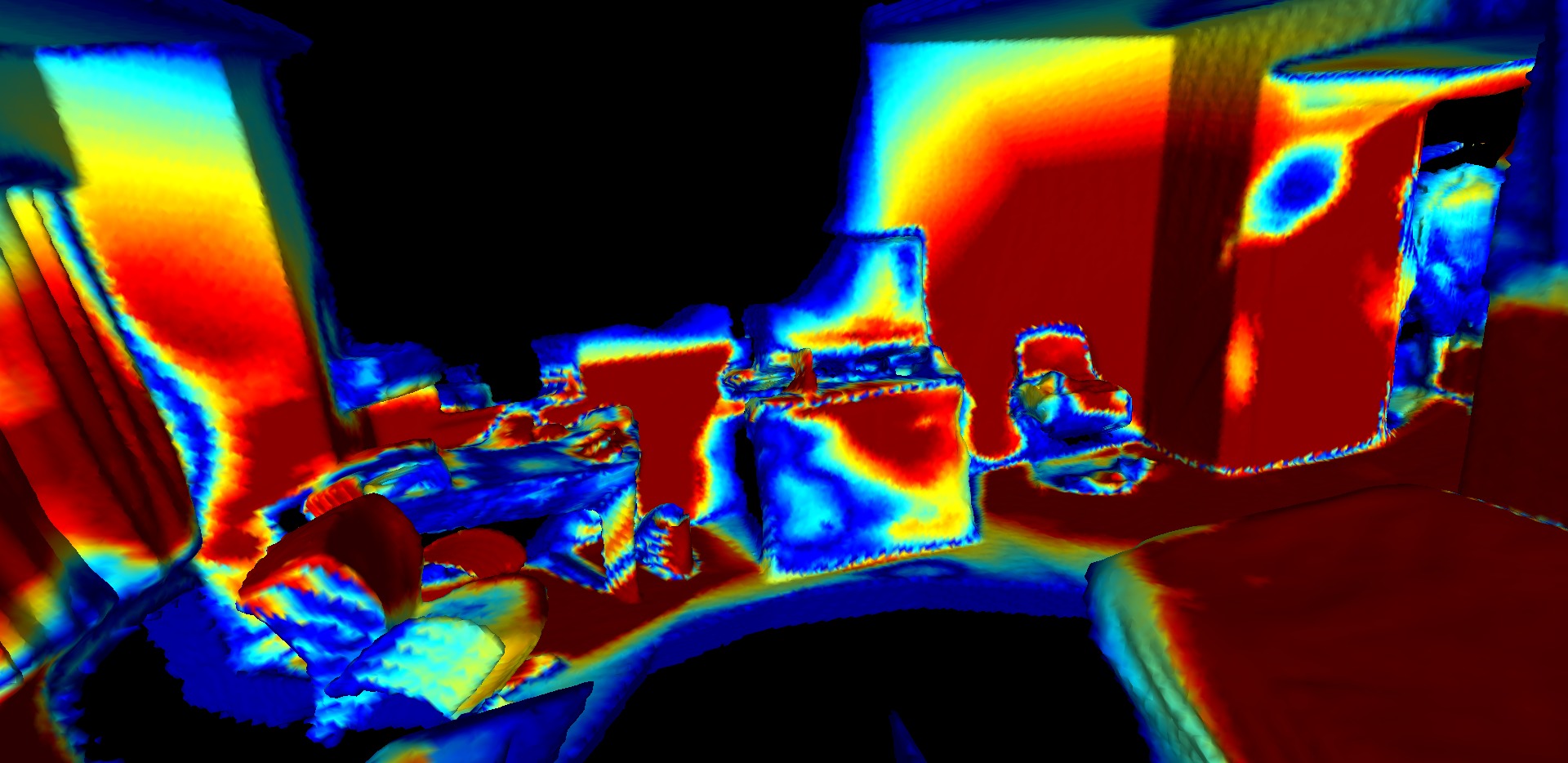}
	\end{subfigure}
	~
	\begin{subfigure}[b]{0.45\linewidth}
		\includegraphics[width=\textwidth]{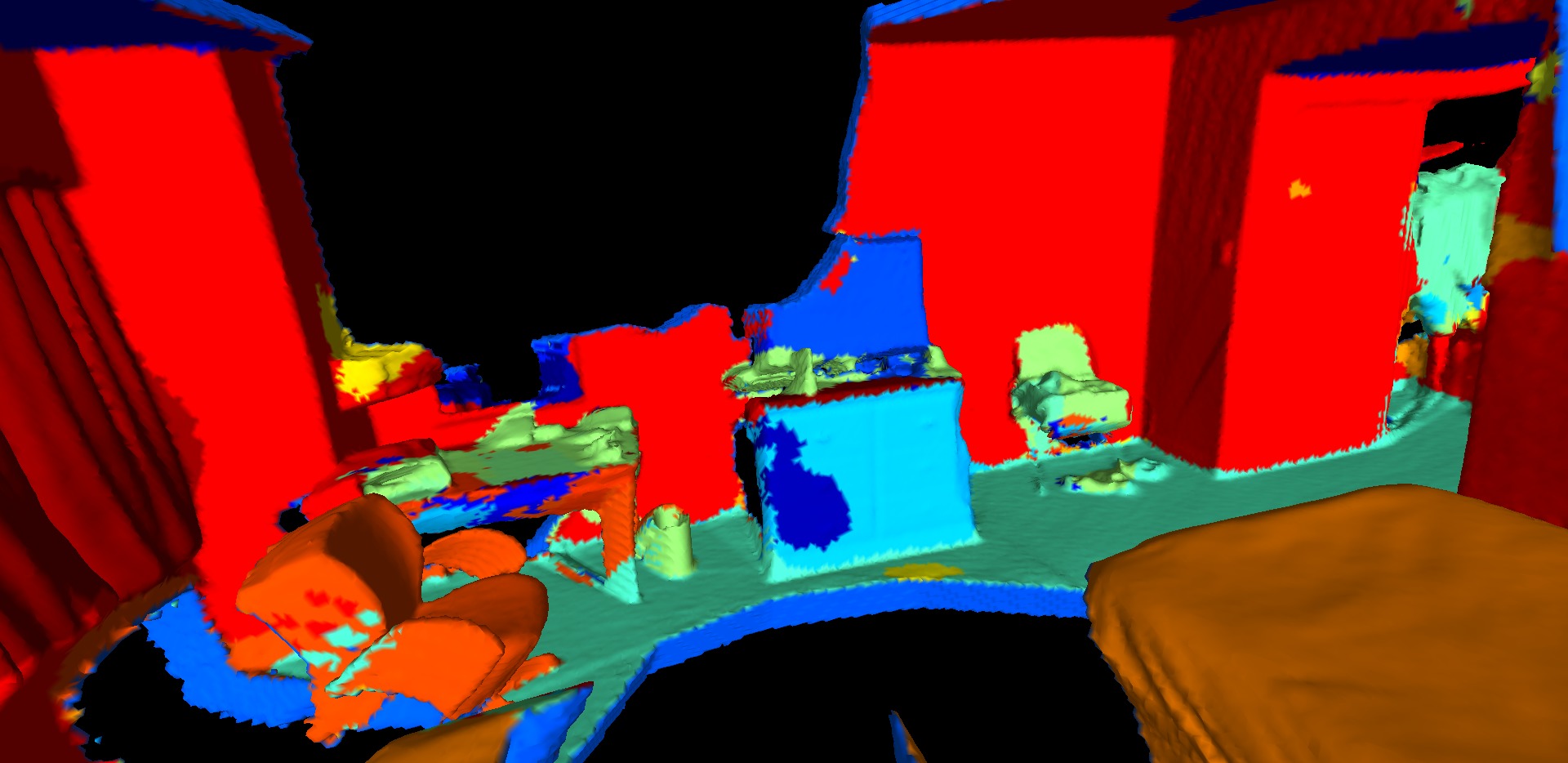}
	\end{subfigure}
	\caption{A second view from the hotel\_umd sequence. Images are ordered as in \autoref{fig:eval_hotel_view_1}.}
	\label{fig:eval_hotel_view_2}
\end{figure*}

\begin{figure*}[htb]
	\centering
	\begin{subfigure}[b]{0.45\linewidth}
		\includegraphics[width=\textwidth]{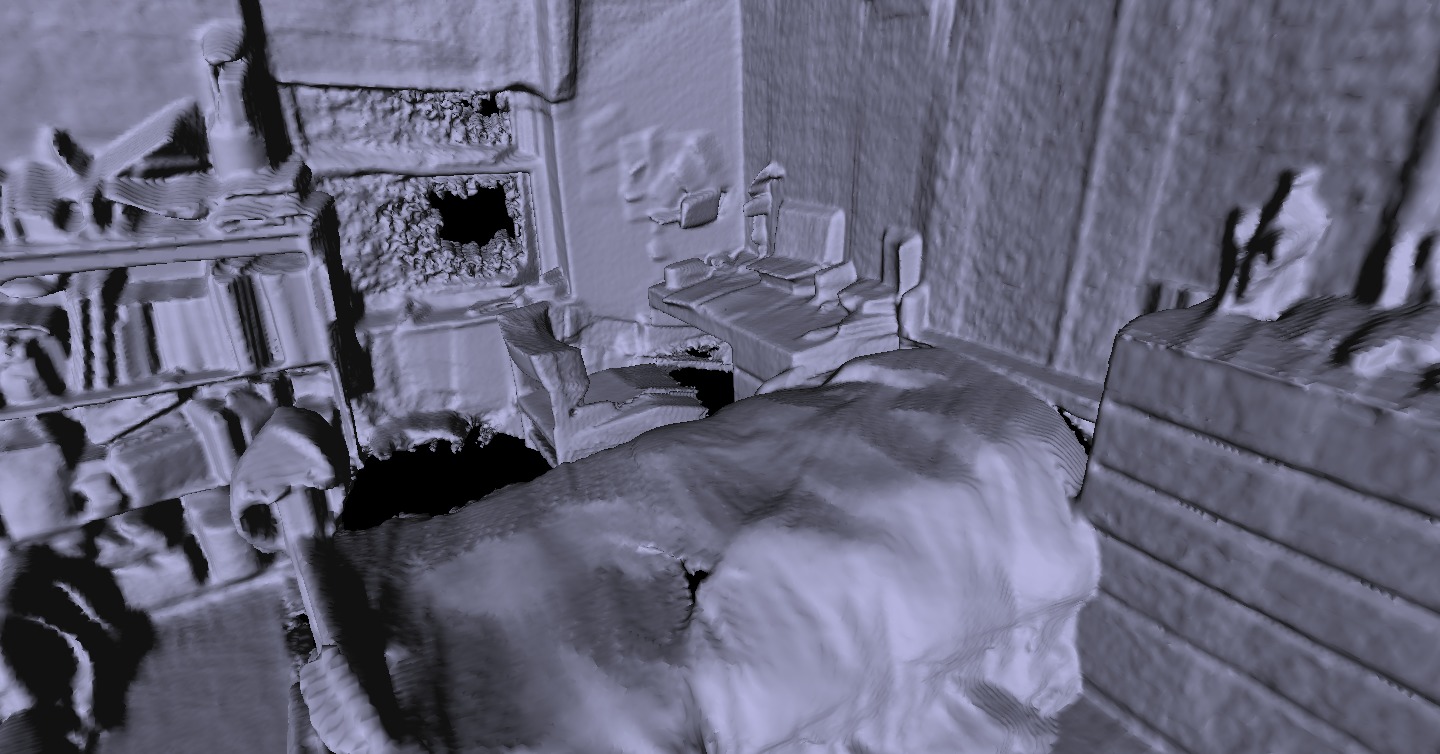}
	\end{subfigure}
	~
	\begin{subfigure}[b]{0.45\linewidth}
		\includegraphics[width=\textwidth]{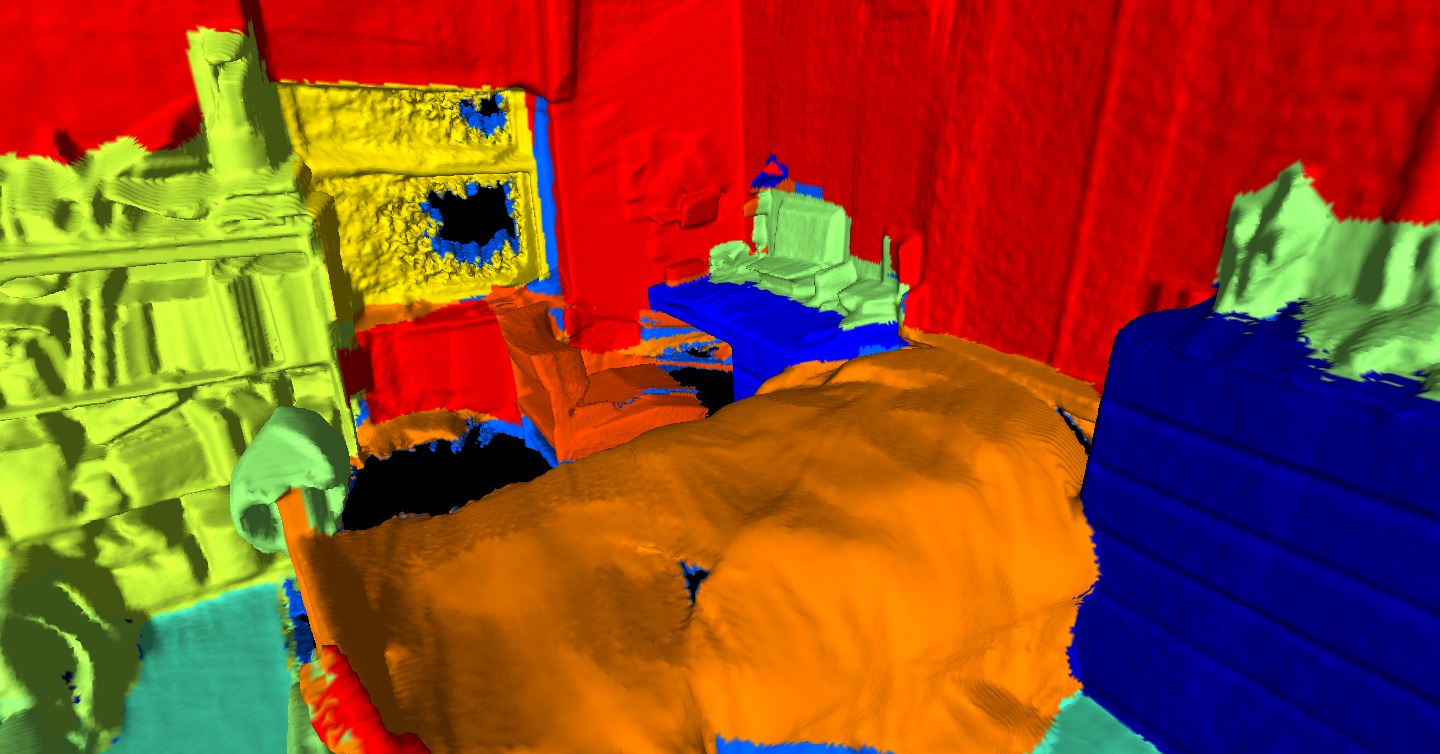}
	\end{subfigure}
	\\
	\begin{subfigure}[b]{0.45\linewidth}
		\includegraphics[width=\textwidth]{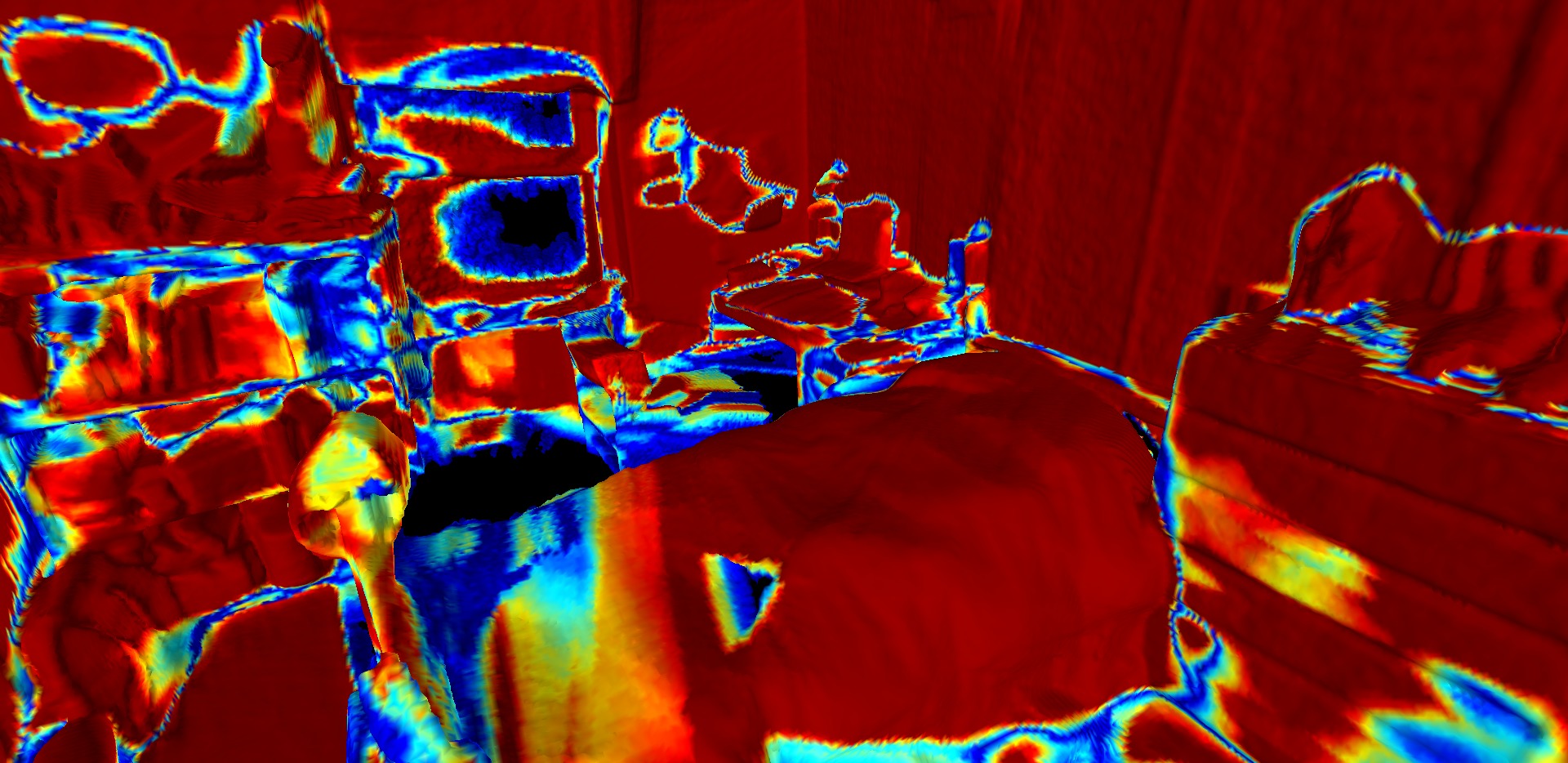}
	\end{subfigure}
	~
	\begin{subfigure}[b]{0.45\linewidth}
		\includegraphics[width=\textwidth]{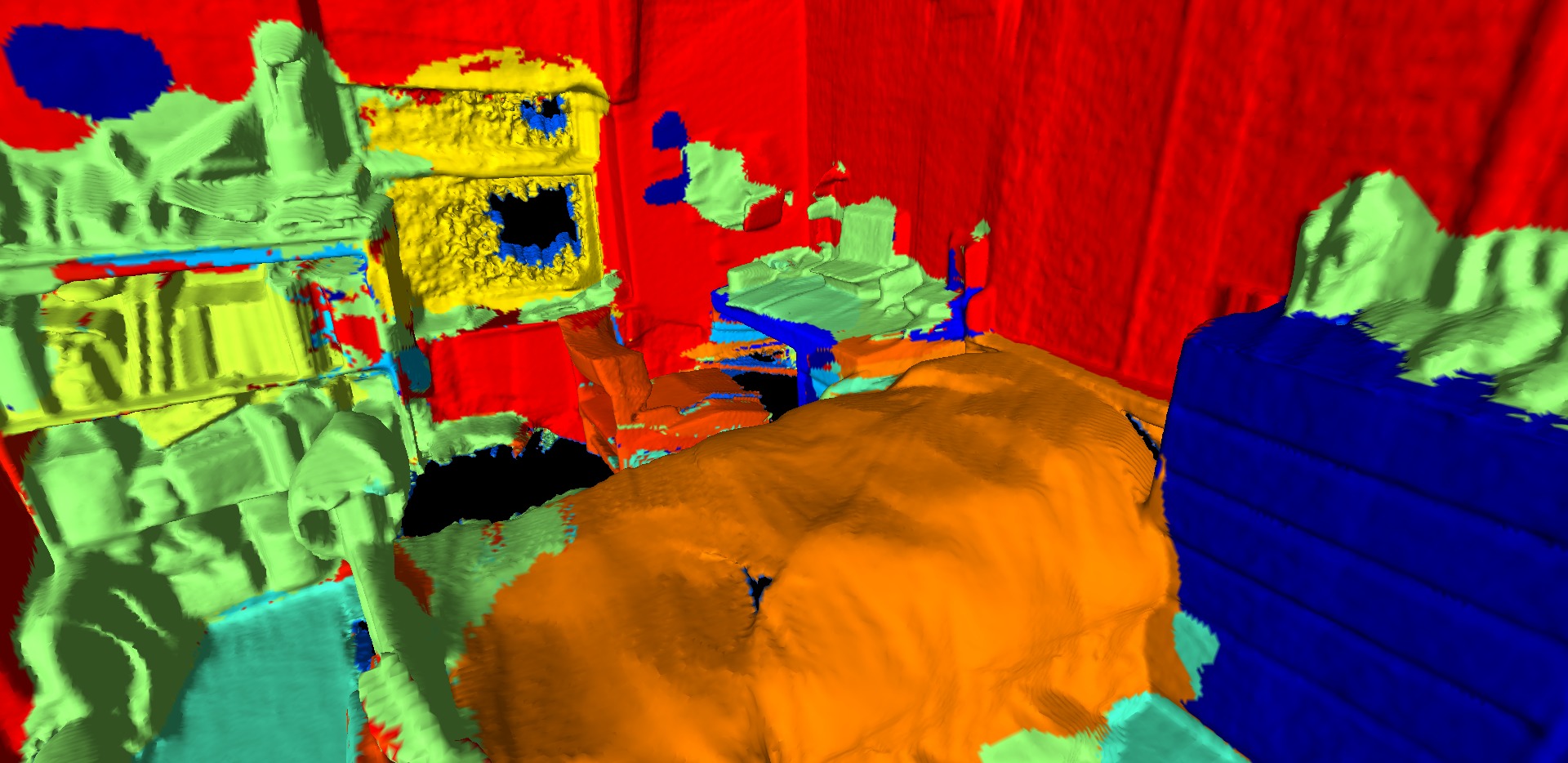}
	\end{subfigure}
	\caption{A view from the mit\_dorm\_next\_sj sequence. Once again, images are ordered as in \autoref{fig:eval_hotel_view_1}.}
	\label{fig:eval_dorm_view_1}
\end{figure*}

Figures~\ref{fig:eval_hotel_view_1},~\ref{fig:eval_hotel_view_2} and~\ref{fig:eval_dorm_view_1} show views from the semantically labeled surfaces obtained by processing some of the sequences of the Sun3D dataset. 
To allow for better comparative assessment of the performance achievable in real settings, in each Figure we report both the reconstruction obtained by feeding our algorithm with  ground truth labels and with the output from the CNN mentioned above.
Identifiers from the Sun3D dataset (greater in number than the 40 categories detected by the algorithm in~\cite{Long2015}) have been manually mapped onto the associated categories in order to have the same objects in both datasets identified by the same numerical identifier (the Sun3D dataset for example, for the hotel sequence, defines four different pillow objects; we mapped all such identifiers to the single ``pillow'' category).
Based on the comparison to the ground-truth reconstructions, it can be observed  that the majority of the labeled regions are consistently and correctly identified by the real algorithm and that, where labeling errors have been made, the associated confidence provided by the proposed label integration technique is likely low (such as in the TV stand in \autoref{fig:eval_hotel_view_2} or on the bed in \autoref{fig:eval_dorm_view_1}).
In the supplementary material we provide a video depicting fully labeled volumes for the two Sun3D sequences ``hotel\_umd'' and ``mit\_dorm\_next\_sj''. 
In the video we show the output of our algorithm when feeding it with manually annotated images and per-pixel categories provided by the CNN.

\begin{table}
	\centering
	\begin{tabular*}{0.85\textwidth}{@{\extracolsep{\fill}} lrrrr}
	\toprule
	& \multicolumn{2}{c}{Per frame error rate (\%)} & \multirow{2}{*}{Volumetric error rate (\%)} \\
	\cmidrule{2-3}
	Sequence         & Average & Std. Dev. & \\
	\midrule
	hotel\_umd          & 34.9 & 22.6 & 24.1 \\
	mit\_dorm\_next\_sj & 26.1 & 15.7 & 19.3 \\
	\bottomrule\\
	\end{tabular*}
	\caption{Volumetric vs. per-frame labeling. The left side of the table reports the error rates yielded by CNN  proposed in~\cite{Long2015} on the individual frames of the two Sun3D sequences considered throughout this paper. The rightmost column shows the percentage of voxels wrongly labeled by our volumtric label integration method. }
	\label{tbl:semantic_results}
\end{table}

Eventually, in \autoref{tbl:semantic_results} we assess the benefits brought in by our volumetric label integration technique with respect to per-frame labeling in real settings, i.e. when deploying a real semantic labeling algorithm such as the CNN proposed in~\cite{Long2015}.
The first part of the table reports per-frame semantic labeling error rates: this metric is computed for each frame using the ground-truth labels provided with the Sun3D dataset so to divide the number of erroneously labeled pixels by the total number of labeled pixels. We show average per-frame error rate and associated standard deviation for the two considered sequences. 
Then, the rightmost column
displays the volumetric error rate, i.e. the percentage of erroneously labeled surface voxels in the final reconstruction of the 3D volume (the same metric as in \autoref{fig:eval_noise}). 
Thus, the results in \autoref{tbl:semantic_results} vouch how the proposed label integration technique can handle effectively varying and large per-frame labeling errors
%(e.g. ranging from less than 10\% to more than 80\%) 
so to provide a significantly more accurate semantic segmentation of the reconstructed environment. It is also worth pointing out that the volumetric error rates reported in \autoref{tbl:semantic_results} turn out higher than those yielded by synthetic label noise (\autoref{fig:eval_noise}) due to the diverse kinds of labeling errors. Indeed, while in the experiment dealing with synthetic noise  each pixel has a uniform and independent probability to be assigned to a wrong category, in real settings it is more likely that large connected image regions get labeled wrongly due to the spatial smoothness constraints enforced  by real semantic labeling algorithms, such as e.g. the CNN deployed in our experiments. 

%------------------------------------------------------------------------
\section{Final remarks}
We have described the first approach to bridge the gap between semantic segmentation and dense surface mapping and tracking, so as to attain, peculiarly, a semantically labeled dense reconstruction of the environment explored by a moving RGB-D sensor. % In this work we proposed a technique to integrate the output of a generic semantic labeling algorithm into a volume-based environment map produced by a SLAM algorithm.
We have demonstrated its robustness by introduction of significant noise in the labeling fed as input as well as its effectiveness by comparing the labeled surfaces achievable by ground truth semantic data to those obtained by deploying a state of the art semantic segmentation algorithm.

Our goal is to provide a tool usable alongside any kind of semantic perception algorithm in order to incrementally gather  high-level knowledge on the environment and store it within the map  itself. 
We also plan to deploy such volumetric semantic data to improve the camera tracking algorithm by exploiting semantic cues together with geometric information to align the current view to the surface embedded into the TSDF volume.
Additionally, it will be feasible to raycast in real time a category and confidence bitmap using the data stored into the volume.
This will allow the user to obtain a continuous stream of semantically labeled frames, possibly interacting with the system while mapping the space so to either linger on low-confidence regions or even correct or improve the acquired semantic information.

%------------------------------------------------------------------------

\bibliographystyle{splncs03}
%\bibliography{library}

\end{document}